\documentclass{article}

\usepackage[main, final]{neurips_2026}

\PassOptionsToPackage{numbers, compress}{natbib}
\usepackage[utf8]{inputenc} 
\usepackage[T1]{fontenc}    
\usepackage{hyperref}       
\usepackage{url}            
\usepackage{booktabs}       
\usepackage{amsfonts}       
\usepackage{nicefrac}       
\usepackage{microtype}      
\usepackage{xcolor}         
\usepackage{algorithm}
\usepackage{amsmath} 
\usepackage{amsthm}
\usepackage{algorithm}
\usepackage{algorithmic}
\usepackage{multirow}
\usepackage{multicol}
\usepackage{subfigure}
\usepackage{subcaption}
\usepackage{colortbl}
\usepackage{enumitem}
\usepackage{wrapfig}
\usepackage{graphicx} 
\usepackage{booktabs}
\usepackage{multirow}
\title{D-GAP: Improving Out-of-Domain Robustness via Dataset-Agnostic and Gradient-Guided Augmentation in Amplitude and Pixel Spaces
}

\author{%
  Ruoqi Wang \\
  Institute of AI for Industries, Chinese Academy of Sciences\\
  The Hong Kong University of Science and Technology (Guangzhou)\\
  \texttt{rqwang@iaii.ac.cn}
  \And
  Haitao Wang \\
  Sun Yat-sen University\\
  \texttt{wanght76@mail.sysu.edu.cn}
  \And
  Shaojie Guo \\
  East China Normal University\\
  \texttt{52275904013@stu.ecnu.edu.cn}
  \And
  Qiong Luo\thanks{Corresponding author.} \\
  The Hong Kong University of Science and Technology\\
  The Hong Kong University of Science and Technology (Guangzhou)\\
  \texttt{luo@cse.ust.hk}
}

\begin{document}

\maketitle

\begin{abstract}
Out-of-domain (OOD) robustness is challenging to achieve in real-world computer vision, especially in unsupervised domain adaptation scenarios, where shifts in image background, style, and acquisition instruments often degrade model performance.  Generic augmentations show inconsistent gains under such shifts, whereas dataset-specific augmentations require expert knowledge and prior analysis. Moreover, prior studies show that neural networks adapt poorly to domain shifts because they exhibit a learning bias to domain-specific frequency components. Perturbing frequency values can mitigate such bias but overlooks pixel-level details, leading to suboptimal performance.  
To address these limitations, we propose \textbf{D-GAP}, a \textbf{D}ataset-agnostic and \textbf{G}radient-guided augmentation method for the \textbf{A}mplitude spectrum (in frequency space) and the \textbf{P}ixel values. 
Unlike conventional handcrafted augmentations, D-GAP computes sensitivity maps in the frequency space from task gradients, which reflect how strongly the deep models respond to different frequency components, and uses the maps to adaptively interpolate amplitudes between source and target samples. 
We further propose a dual-space augmentation that jointly controls spectral bias and spatial fidelity by introducing a complementary pixel-space blending branch.
This way, D-GAP turns augmentation from fixed, random, or manually designed perturbation into a model-response-adaptive intervention.
Extensive experimental results show that the proposed method consistently outperforms both generic and dataset-specific domain adaptation methods, improving average OOD performance by +5.3\% on four real-world datasets and +1.9\% on three benchmark datasets. Code is available at https://github.com/RapidsAtHKUST/D-GAP.
\end{abstract}

\section{Introduction}
\label{sec:intro}
In real-world applications, the distributions in the training datasets often differ from those in other deployment environments because of the shifts in background and acquisition instruments \cite{koh2021wilds, qu2024connect}. In such out-of-distribution (OOD) settings, models trained on source domains always degrade when applied to unlabeled target domains. For example, in wildlife monitoring, ecologists use models to classify species of camera trap images, but these models often show significant accuracy drops when applied to new locations \cite{beery2018recognition}. Annotated data are usually available only for a limited subset of cameras, which may not contain the diversity of environments in the unlabeled target domains. In this paper, we focus on improving OOD robustness in domain adaptation settings, where we have labeled data from source domains and unlabeled data from target domains. 

Data augmentation is a common strategy for improving OOD robustness of deep-learning models, but designing effective augmentations remains challenging. Generic augmentations of images (e.g., RandAugment \cite{cubuk2020randaugment}, CutMix \cite{yun2019cutmix}, FACT \cite{xu2023fourier}, SAM \cite{xu2023semantic}) improves the performance across datasets to various degrees \cite{gulrajanisearch, wilesfine, hendrycks2021many}, whereas domain-invariance methods \cite{yan2020improve, zhou2020deep, ilse2021selecting, gulrajanisearch, yao2022improving, wong2025approximate, su2024enhanced} yield limited improvements, especially on real-world domain shifts \cite{gao2023out}. Although some dataset-specific augmentation methods \cite{gao2023out, qu2024connect} can improve OOD performance, they often need expert knowledge and prior analysis of the dataset, which makes these methods hard to scale or apply to new datasets.
These limitations highlight the need for more general, dataset-agnostic methods to improve OOD robustness across diverse real-world settings.

Prior studies show that neural networks tend to learn biasedly from different frequency components depending on the dataset’s frequency characteristics \cite{pinson2023linear, he2024towards, wang2023neural}. Such spectral bias causes the model to overfit domain-specific frequency components, leading to poor generalization when the target domain has different frequency distributions. As such, we consider that this bias can be reduced by perturbing the the frequency distributions. 
However, existing frequency-based augmentation methods \cite{xu2023semantic, xu2023fourier, yang2020fda49, yang2020phase48, yucel2023hybridaugment, nam2024fsda, lee2024domain, salehnia2025frequency, huang2021fsdr} typically perturb frequency components using predefined, manually constrained rules, or cannot integrate information from unlabeled target domain. This lack of model-response awareness limits their performance in domain adaptation tasks.

\begin{figure}[t]
\centerline{\hspace*{8mm}\includegraphics[height=2.2cm]{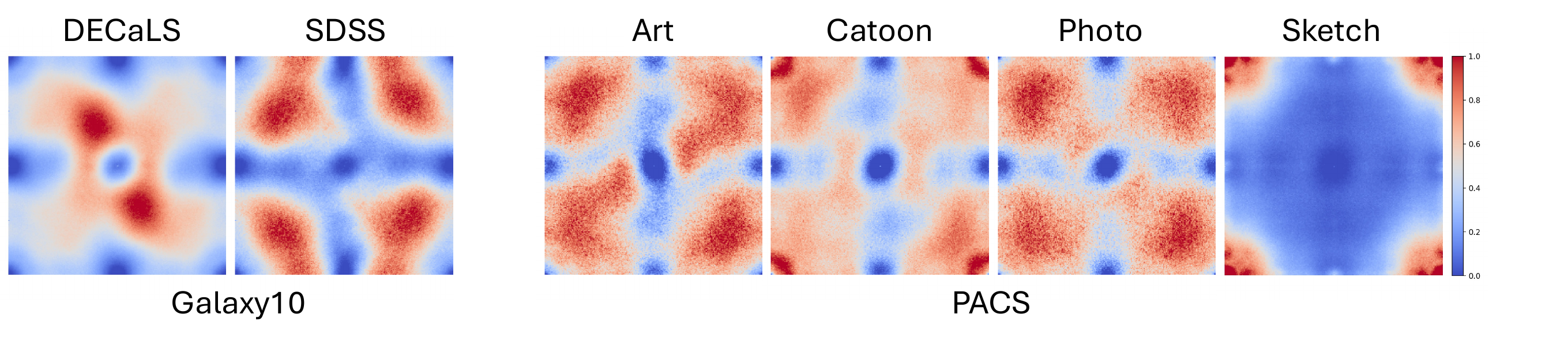}}
\caption{Average amplitude sensitivity maps of ResNet50 trained across domains.}
\label{sensitivity_maps}
\end{figure}

To address these problems, we propose \textbf{D-GAP}, a domain adaptation framework that improves OOD robustness by introducing targeted augmentations in both frequency and pixel spaces. Unlike previous targeted augmentations with manually designed, dataset-specific rules, D-GAP automatically adapts to domain shifts from data and models, without the need for expert knowledge or prior analysis. 
Specifically, given two images from different domains, in the frequency space, we introduce a gradient-adaptive mechanism that computes amplitude sensitivity maps from the task gradients with respect to the amplitude spectrum (as shown in Fig.~\ref{sensitivity_maps}). These sensitivity maps represent how strongly the models respond to different amplitude components. Instead of interpolating two spectra with a predefined or random ratio, D-GAP adjusts the interpolation strength at each frequency according to its sensitivity value: more sensitive amplitude components are more strongly mixed with target-domain amplitude spectrum, while less sensitive ones are more preserved. This mechanism enables the models to suppress domain-specific spectral learning bias.
This gradient-guided interpolation creates frequency-space augmentations that adaptively keep the main content of the source image while adding features from target domains. However, reconstructed results of amplitude blending sometimes introduce artifacts and blurring. As such, we apply a pixel value blending to add complementary spatial pixel-level information. The final augmented image is then obtained by blending these two results from both the frequency space and the pixel space.

We firstly evaluate our method on four real-world datasets: 
Compared to other generic methods, D-GAP achieves improvements in OOD performance, including +2.1\% gains on iWildCam \cite{beery2021iwildcam, sagawaextending} (wildlife recognition), +4.2\% on Camelyon17 \cite{bandi2018detection, sagawaextending} (tumor detection), +5.6\% on BirdCalls \cite{joly2022overview, gao2023out} (birdcall recognition), and +9.3\% on Galaxy10 \cite{Galaxy10} (galaxy morphology classification). Also, our method surpasses dataset-specific augmentation strategies tailored for different datasets. Additionally, D-GAP also makes average gains of +1.9\% accuracy on standard benchmark datasets (PACS \cite{li2017deeper}, Office-Home \cite{venkateswara2017deep}, and Digits-DG \cite{zhou2020deep}), showing its improvement across diverse distribution shifts. We further analyze the generalizability of D-GAP on various backbone networks and its effectiveness of connecting same-class cross-domain samples while separating different-class examples.

Our work makes the following contributions: 
(1) We propose a model-response-adaptive augmentation for OOD robustness in domain adaptation, where the perturbation intensity is determined by the amplitude sensitivity of models rather than by manually designed rules.
(2) We propose Gradient-guided Amplitude Mix, which computes gradient-guided amplitude sensitivity maps and performs cross-domain interpolation. 
(3) We introduce a dual-space augmentation design that couples gradient-guided spectral intervention with pixel-space detail compensation.
(4) We achieve SOTA results across multiple backbones on both real-world and general benchmark datasets.

\section{Background and Formulation}



\subsection{Feature Decomposition}
\label{Feature Decomposition}
To better understand how models perform under domain shifts, we use the feature decomposition framework \cite{gao2023out}, which categorizes input features based on their dependence on the label and domain. Specifically, input features can be decomposed into four types based on their dependence on the label and the domain: 
(1) features that are \textit{label-dependent} and \textit{domain-independent} ($x_\text{obj}$); 
(2) \textit{label-dependent} and \textit{domain-dependent} features ($x_\text{d:robust}$); 
(3) \textit{label-independent} and \textit{domain-dependent} features ($x_\text{d:spu}$); and 
(4) \textit{label-independent} and \textit{domain-independent} features ($x_\text{noise}$).
 We formalize these relationships via their dependence with the label $y$ and domain $d$:
\[
x_\text{obj},\; x_\text{d:robust} \not\!\perp\!\!\!\perp\; y, \quad
x_\text{noise},\; x_\text{d:spu} \perp\!\!\!\perp\; y, \quad 
x_\text{d:robust},\; x_\text{d:spu} \not\!\perp\!\!\!\perp\; d, \quad
x_\text{obj},\; x_\text{noise} \perp\!\!\!\perp\; d.
\]
Models should focus on both $x_\text{obj}$ and $x_\text{d:robust}$ while avoiding reliance on spurious or noisy features \cite{gao2023out}. We later analyse these relationships using a connectivity-based framework (Section \ref{Sec:Connectivity} and \ref{Empirical Evaluations of Connectivity}) to evaluate how augmentations influence these label/domain dependent features.

\begin{figure*}[t]
\centerline{\includegraphics[height=5.5cm]{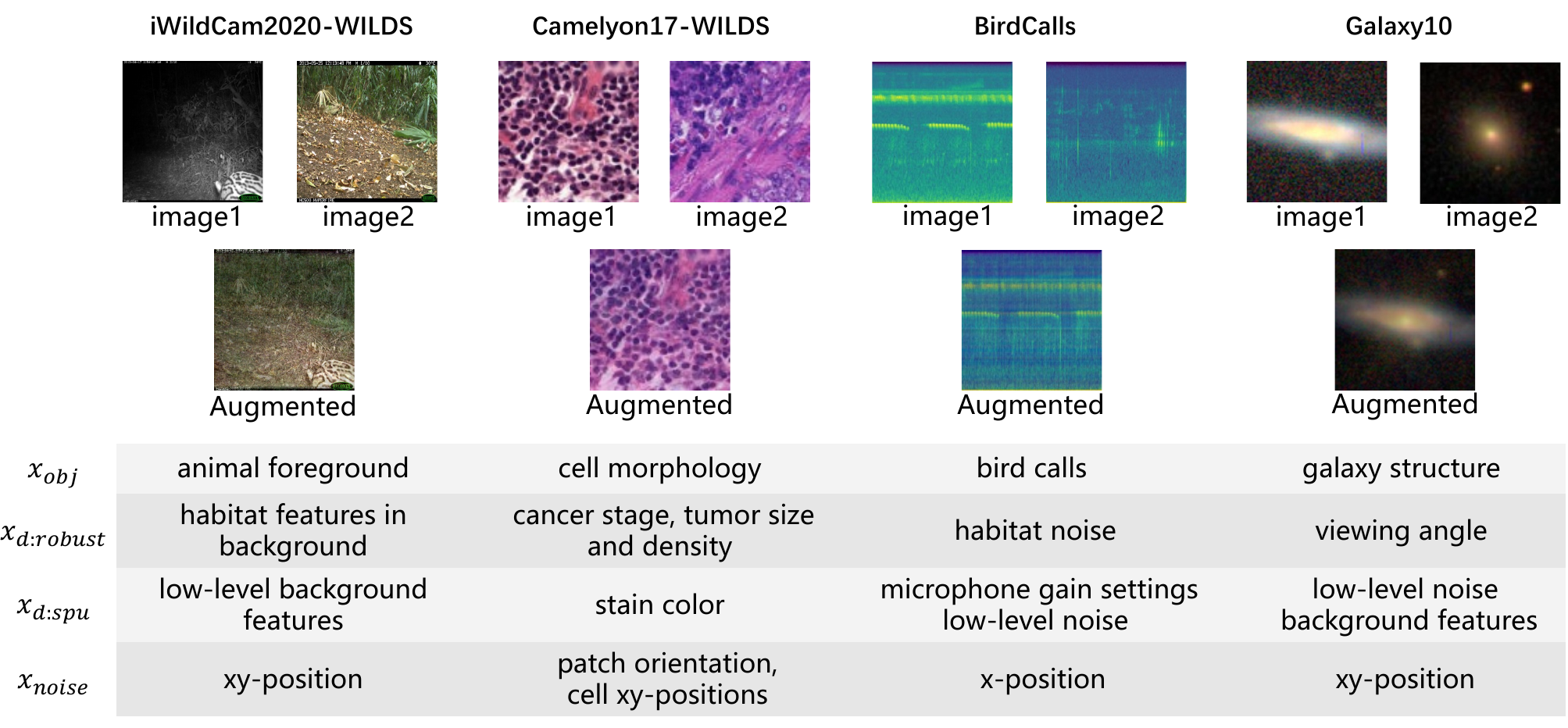}}
\caption{
\textbf{Feature decomposition and augmentation examples.} 
\textit{Top:} For each dataset (iWildCam, Camelyon17, BirdCalls, Galaxy10), we show a source image, its corresponding augmented image generated by our method.
\textit{Bottom:} Based on the decomposition framework \cite{shen2022connect}, we annotate representative features in the datasets across $x_{\text{obj}}$, $x_{d\text{:robust}}$, $x_{d\text{:spu}}$, $x_{\text{noise}}$. 
Our method effectively randomizes $x_{d\text{:spu}}$, varies $x_{d\text{:robust}}$ while preserving or  $x_{\text{obj}}$.
}
\label{4x}
\end{figure*}

\subsection{Datasets}

\textbf{Real-World Datasets.}
As shown in Fig. \ref{4x}, we test our method on four real-world datasets in this paper: 
\textbf{iWildCam} \cite{beery2021iwildcam,koh2021wilds} -- wildlife species classification with location-related background shifts.
\textbf{Camelyon17} \cite{bandi2018detection,koh2021wilds} -- tumor identification across hospitals with stain and density variations. 
\textbf{BirdCalls} \cite{joly2022overview,gao2023out} -- bird species recognition under habitat and microphone domain noise.
\textbf{Galaxy10} \cite{Galaxy10} -- galaxy morphology classification across telescopes differing in resolution and color profile.
Each dataset presents distinct domain shifts such as environmental background, staining style, recording device, or instrument differences.
More dataset details are provided in Appendix Sec \ref{Dataset Details}.

\textbf{Additional Benchmark Datasets.}
In addition to the four real-world datasets used throughout this paper, we further evaluate our method
on three standard benchmarks: \textbf{PACS}~\cite{li2017deeper},
\textbf{Office-Home}~\cite{venkateswara2017deep}, and \textbf{Digits-DG}~\cite{zhou2020deep}. Compared to our four real-world datasets that cover diverse real-world scenario and deployment-related shifts, these benchmark datasets focus on visual recognition with well-studied style variations.  
More details are provided in Appendix Sec. \ref{Dataset Details}.

\subsection{Targeted Augmentations}
Some dataset-specific augmentation methods \cite{gao2023out, qu2024connect} can improve OOD performance on real-world applications. However, they often need expert knowledge and prior analysis of the dataset, which makes these methods hard to scale or apply to new datasets.
For instance, in wildlife recognition, some methods randomize the background of camera trap images to mitigate spurious correlations between results and location-specific terrains or lighting conditions. Such a target augmentation helps the model to focus on the animal foreground. However, it depends on the segmentation of foreground and background, restricting its application to datasets without foreground annotations. Similarly, in tumor identification from histopathology slides, Stain Color Jitter can deal with stain color variations across hospitals \cite{tellez2018whole}. It perturbs images within the hematoxylin and eosin staining features to overcome color-related domain biases, while preserving morphological features \cite{gao2023out}. However, it requires expert knowledge of staining chemistry and color editing. These examples show that while dataset-specific augmentations can be effective, they are difficult to generalize.

\subsection{Connectivity}
\label{Sec:Connectivity}

\begin{wrapfigure}[15]{r}{0.55\textwidth}
\includegraphics[width=7.8cm]{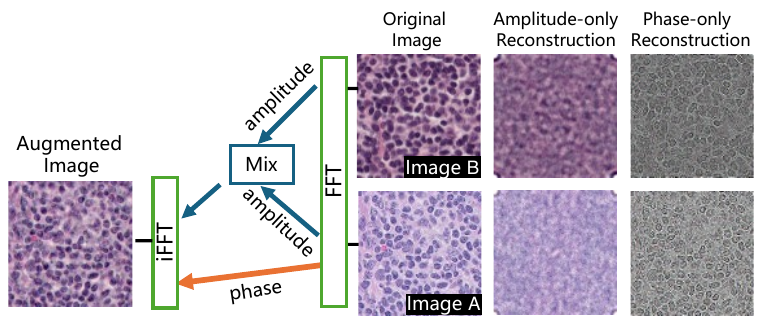}
\caption{Amplitude-only and phase-only reconstructions of two images. The amplitudes of Image A and B are first mixed, and the mixed amplitude is combined with the phase of Image A to generate an augmentation.}
\label{frequency}
\end{wrapfigure}

\begin{figure*}[t]
\centerline{\includegraphics[height=4.8cm]{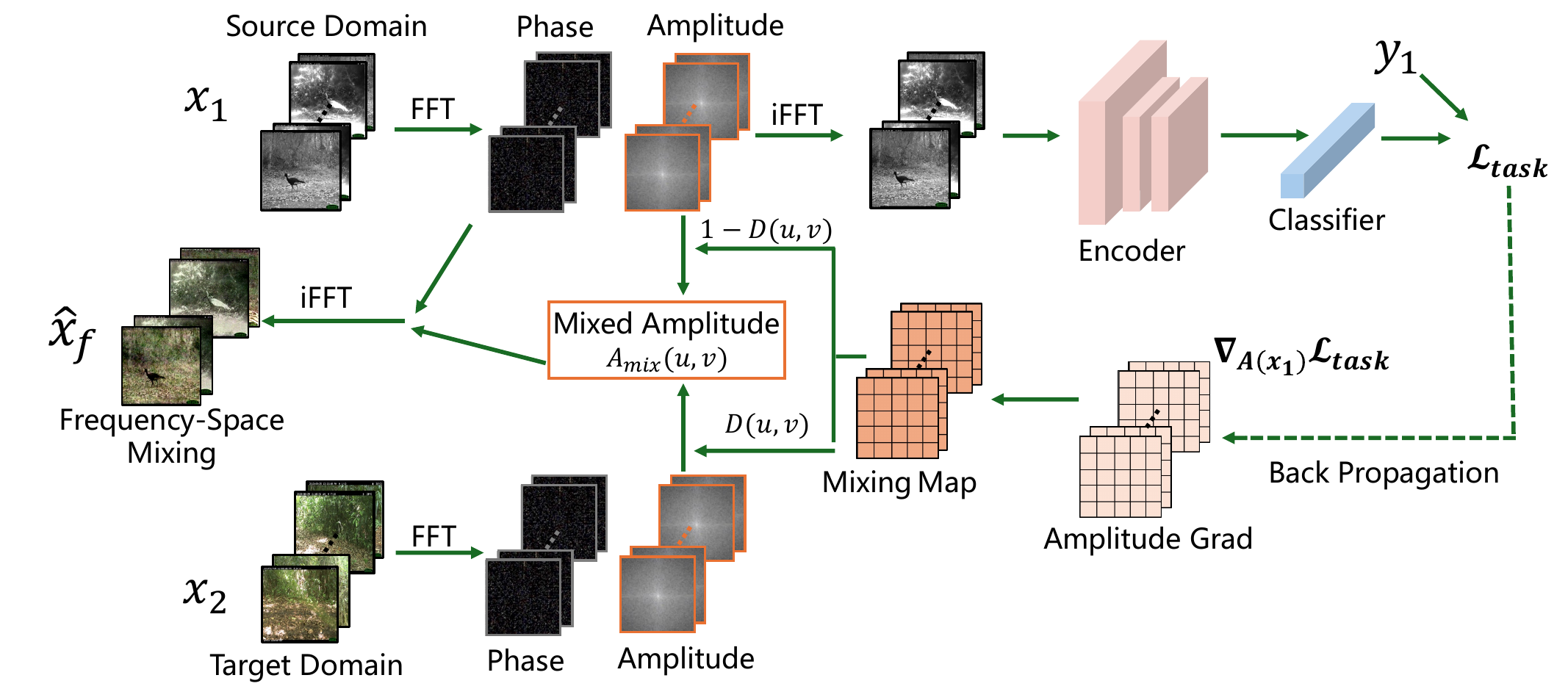}}
\caption{Overview of the Gradient-guided Amplitude Mix procedure. Given a source image $x_1$ and a target-domain image $x_2$, we compute the Sensitivity Map $G(u,v)$ and the Mixing Map $D(u,v)$. The mixed amplitude is then combined with the phase of $x_1$. }
\label{ouraug}
\end{figure*}

To better understand how data augmentations impact cross-domain generalization, we adopt the concept of \textit{connectivity} \cite{shen2022connect}. 
We define connectivity between a class-domain pair $((y_1, d_1), (y_2, d_2))$ under four conditions: 
(1) $\rho : y_1 = y_2,\ d_1 = d_2. $
(2) $\alpha : y_1 = y_2,\ d_1 \ne d_2. $
(3) $\beta : y_1 \ne y_2,\ d_1 = d_2. $
(4) $\gamma : y_1 \ne y_2,\ d_1 \ne d_2. $
Here, each of the $\rho$, $\alpha$, $\beta$, $\gamma$ values indicates how “connected” a pair is. Shen et al. \cite{shen2022connect} show that the ratios $\frac{\alpha}{\gamma}$ and $\frac{\beta}{\gamma}$ exhibit strong empirical correlation with OOD accuracy. In this work, we follow this approach to show how our augmentation changes the connectivity of the data. More details on the computation of connectivity are presented in Section \ref{Empirical Evaluations of Connectivity} and the Appendix Sec. \ref{appendix_connectivity}.
To improve OOD robustness, the model must randomize features that depend on specific domains while preserving task-relevant information.
Specifically, in feature decomposition~\cite{shen2022connect}, we randomize domain-dependent features ($x_{d\text{:spu}}$) and keep label-relevant ones ($x_{\text{obj}}$, $x_{d\text{:robust}}$).
As shown in Fig.~\ref{frequency}, we observe that amplitude mixing in frequency domain (Details of the mixing strategy are provided in Section \ref{Method}.) preserves task-relevant semantic structure \cite{xu2023fourier} ($x_{\text{obj}}$ and $x_{d\text{:robust}}$), while effectively randomizing $x_{d\text{:spu}}$. However, solely frequency mixing may introduce artifacts and blurring (as examples in Fig. \ref{frequency} and \ref{ouraug}), influencing the extraction of detailed features.

\begin{wraptable}{r}{6.3cm}
\centering
\footnotesize
\caption{Connectivity values on iWildCam.}
\begin{tabular}{l|ccc|cc}
\toprule
\textbf{Aug} & $\alpha$ & $\beta$ & $\gamma$ & $\alpha/\gamma$ & $\beta/\gamma$ \\
\midrule
w/o & 0.007 & 0.008 & 0.021 & 0.33 & 0.38 \\
Ours     & 0.212 & 0.030 & 0.051 & 4.16 & 0.59 \\
\bottomrule
\end{tabular}
\label{connect_test}
\end{wraptable}


To address the problem, we add pixel-space mixing as a complement, which brings back more pixel-level details and perturbs $x_{d\text{:spu}}$. However, it may influence $x_{\text{obj}}$ and $x_{d\text{:robust}}$ as well. Therefore, we then define a tunable ratio (details are in Section \ref{Method}). By combining the augmentation in both frequency and pixel spaces, our goal is to balance the label-relevant and domain-relevant shifts, achieving domain diversity while maintaining semantic identity. This dual-space augmentation enables fine-grained control over domain perturbation intensity and promotes robust feature learning across source and unseen domains. 
We then empirically evaluate our method by comparing the connectivity ratios $\frac{\alpha}{\gamma}$ and $\frac{\beta}{\gamma}$ of the augmented and unaugmented datasets. Results in Table \ref{connect_test} show that augmentation in frequency-pixel spaces can enhance the cross-domain connectivity, achieving domain diversity while maintaining label-relevant identity.

\section{Method}
\label{Method}


In this section, we introduce the technical details of D-GAP, including Fourier transform preliminaries and details of each module in our method, including the Gradient-guided Amplitude Mix, Pixel-Space Mixing, and the training framework.

\vspace{-5pt}
\subsection{D-GAP}
\textbf{Fourier Transform of Images.}
For an image $x$, its Fourier transformation $\mathcal{F}(x)$ is formulated as:
\begin{equation}
\small 
\mathcal{F}(x)(u, v) = \sum_{h=0}^{H-1} \sum_{w=0}^{W-1} x(h, w) e^{-j2\pi\left( \frac{hu}{H} + \frac{wv}{W} \right)},
\end{equation}
where $(u,v)$ are coordinates in the frequency space. $\mathcal{F}^{-1}(x)$ denotes the corresponding inverse Fourier transformation. Both forward and inverse Fourier transforms can be efficiently computed via the Fast Fourier Transform (FFT) algorithm.

The amplitude and phase components of $\mathcal{F}(x)$ are computed respectively as:

\begin{equation}
\small
\begin{aligned}
\mathcal{A}(x)(u,v) = \big[ R^2(x)(u,v) + I^2(x)(u,v) \big]^{1/2},
\mathcal{P}(x)(u,v) = \arctan\!\left(\frac{I(x)(u,v)}{R(x)(u,v)}\right).
\end{aligned}
\end{equation}
where $R(x)$ and $I(x)$ denote the real and imaginary parts of $\mathcal{F}(x)$, respectively. For RGB images, the Fourier transform is applied to each channel separately.



\textbf{Gradient-guided Amplitude Mix. }
Conventional frequency-based augmentations~\cite{xu2023fourier,xu2023semantic} typically mix amplitudes of two images with a random intensity factor.
While such random interpolation introduces diversity, it fails to distinguish how sensitive that the model is to each frequency component. Moreover, the interpolation ratio needs to be manually constrained within a predefined range to avoid excessive distortion.
To overcome these limitations, D-GAP performs a gradient-guided amplitude mix, adaptively adjusting the interpolation strength based on the model’s sensitivity in the frequency space. The process of Gradient-guided Amplitude Mix is shown in Fig. \ref{ouraug}.

Given two images $x_1$ and $x_2$, we aim to augment $x_1$ by injecting the amplitude information of $x_2$ into a square region $\Omega_r$ in the Fourier space.
Let $f$ denote the classifier and $\mathcal{L}_{\text{task}}$ the supervised loss.
We compute the gradient of the loss with respect to the source amplitude:
\begin{equation}
\label{eq3}
\small
G(u,v)=\Bigl|\frac{\partial\,\mathcal{L}_{\text{task}}(f(x_1),y)}{\partial\,A(x_1)(u,v)}\Bigr|,\quad (u,v)\in\Omega_r.
\end{equation}
Here, $G(u,v)$ measures the model’s sensitivity to each frequency, which means how much the task loss would change if the amplitude at a specific frequency $(u,v)$ were perturbed. In this paper, we call $G(u,v)$ \textbf{sensitivity map}. Large gradient values indicate that the model’s prediction heavily depends on that frequency component, implying stronger spectral bias. This allows D-GAP to continuously regulate how much information to mix at each frequency.

Then after normalizing $G$ and applying a sigmoid function,
\begin{equation}
\small
\tilde{G}=\frac{G-\mu(G)}{\sigma(G)+\varepsilon},\quad
D=\mathrm{clip}\big(\mathrm{Sigmoid}(\tilde{G}),d_{\min},d_{\max}\big),
\end{equation}
we obtain a gradient-based \textbf{mixing map} $D(u,v)$ from the sensitivity map $G(u,v)$ that adaptively controls the amplitude interpolation. In our design, higher sensitivity (larger $G(u,v)$) leads to stronger interpolation from the target domain, encouraging the model to replace its frequency-biased components with cross-domain information.
Conversely, D-GAP uses weaker interpolation in lower sensitivity regions to preserve non-biased features. The final adaptive interpolation is computed as:
\begin{equation}
\small
\label{eq:afm}
\begin{aligned}
A_{\text{mix}}(u,v)
= \bigl(1-D(u,v)\bigr)\,A(x_1)(u,v) + D(u,v)\,A(x_2)(u,v),\qquad (u,v)\in\Omega_r .
\end{aligned}
\end{equation}
Finally, the mixed image is reconstructed as
\begin{equation}
\small
\label{eq:recon-afm}
\begin{aligned}
\mathcal{F}_{\text{mix}}(u,v)
= A_{\text{mix}}(u,v)\,e^{-jP(x_1)(u,v)}, 
\hat{x}_f
= \mathcal{F}^{-1}\!\bigl(\mathcal{F}_{\text{mix}}\bigr).
\end{aligned}
\end{equation}
Through this gradient-guided strategy, D-GAP adaptively perturbs spectral values with target domain more strongly where the model exhibits higher frequency bias.
This data-driven, continuous modulation mitigates models' reliance on domain-specific spectral shortcuts, encouraging the model to learn more balanced and generalizable representations across domains.

\textbf{Pixel-Space Mixing:}  
Frequency blending sometimes introduces artifacts and blurring in the resulting augmentations. As a result, in addition to frequency-domain augmentation, we introduce a pixel-space mixing strategy to add more pixel-level details in the augmentation. Specifically, given a source image $x$ and a target image $x'$, we apply pixel-wise blending with ratio $\lambda_1$:
\begin{equation}
\small
\hat{x}_p = (1 - \lambda_1)x_1 +\lambda_1 x_2,
\end{equation}
Then we combine pixel-space and frequency-space augmented images via a second-stage blending:
\begin{equation}
\small
\hat{x} = (1 - \lambda_2)\hat{x}_f + \lambda_2 \hat{x}_p,
\end{equation}
This two-stage fusion allows us to perturb frequency bias while maintaining the semantic content of the original sample and the pixel details.

\subsection{Training Framework}
Our training strategy follows established practices in prior works \cite{qu2024connect, kumar2022fine, xu2023fourier, xu2023semantic}: For \textbf{real-world datasets} (e.g., \textit{iWildCam}, \textit{Camelyon17}, \textit{BirdCalls}, and \textit{Galaxy10}), we use the \textbf{Linear Probing then Fine-Tuning (LP-FT)} strategy~\cite{qu2024connect, kumar2022fine}. 
We first train a linear classifier on frozen pretrained features to stabilize early optimization, then fine-tune both the encoder and classifier using D-GAP augmentations.
This procedure improves OOD robustness by gradually adapting high-level representations to domain-diverse augmentations while preventing early-stage overfitting. For other \textbf{common DG benchmark datasets} (i.e., \textit{PACS}, \textit{Office-Home}, \textit{Digits-DG}), we follow previous works \cite{xu2023fourier, xu2023semantic} to train directly on the pretrained encoder without the LP-FT stage for fair comparisons. 
On these datasets, the pretrained model already provides a stable initialization, and applying D-GAP augmentations directly during training yields efficient adaptation and robust convergence.

\section{Experiments}
\vspace{-5pt}
In this section, we first conduct comparisons between D-GAP and other recent methods on various datasets. We also conduct ablation studies to evaluate the contribution of each module in D-GAP. We then evaluate how our method generalizes beyond backbones. Finally, we conduct a connectivity analysis on empirical evidence of how our method enhances cross-domain feature alignment. Additionally, we present the analysis over hyperparameters including $\Omega_r$ and the blending weights $\lambda_1$, $\lambda_2$ in Appendix Sec. \ref{Analysis over Hyperparameters}.

\begin{figure*}[t]
  \centering
  \begin{subfigure}{}
    \includegraphics[width=11cm]{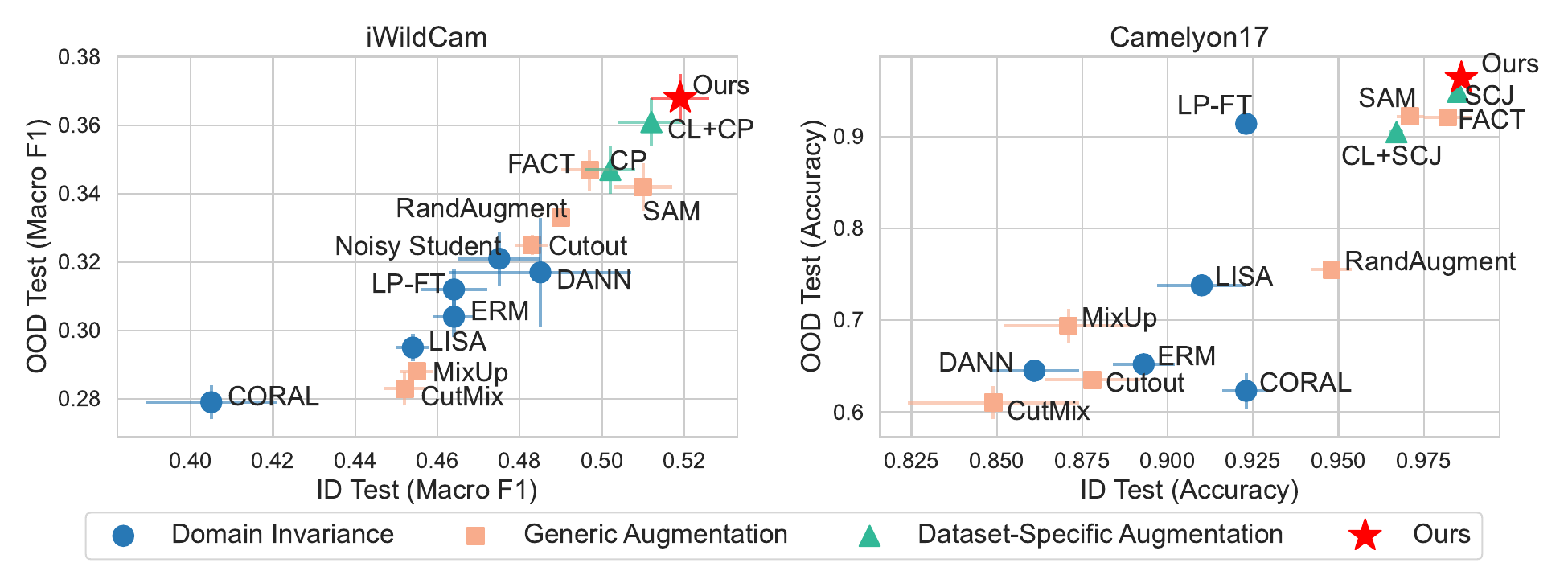} 
  \end{subfigure}
  \\
  \begin{subfigure}{}
    \includegraphics[width=11cm]{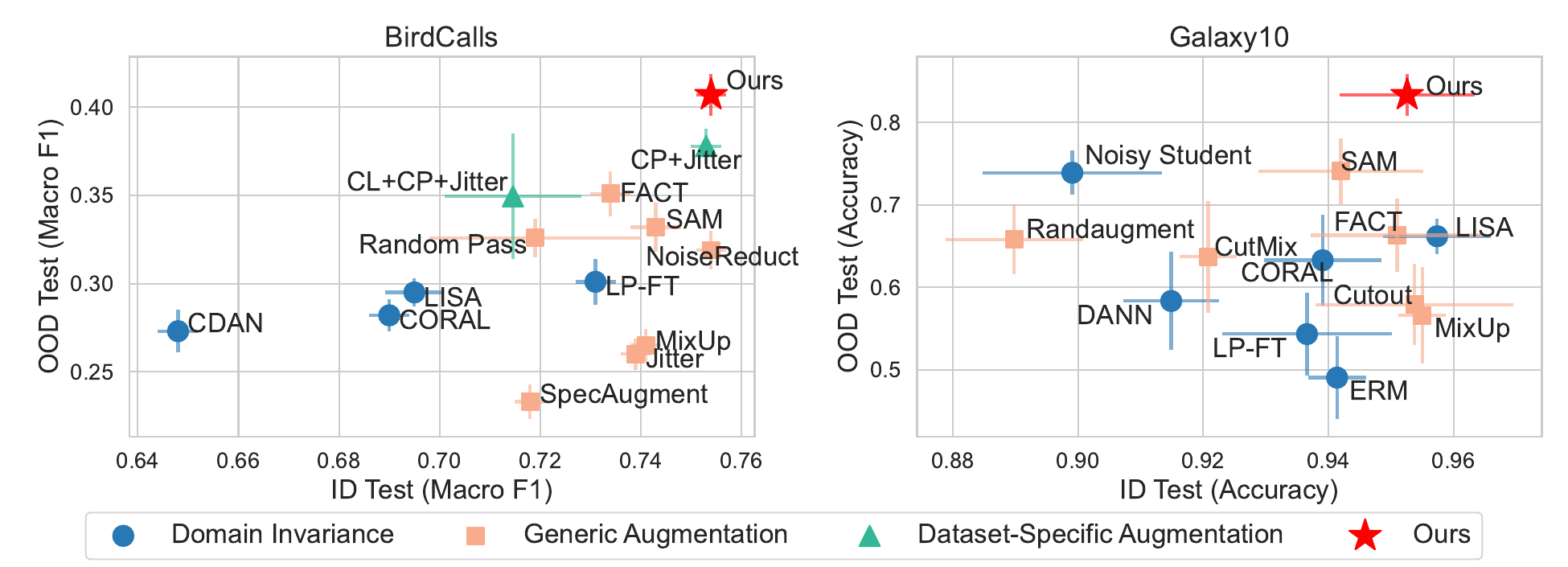}
  \end{subfigure}
  \caption{We plot the in-domain (ID) versus OOD performance across datasets (mean and std). `CL' refers to `Connect Later'. `CP' refers to `Copy Paste'. `SCJ' refers to `Stain Color Jitter'. }
  \label{points}
\end{figure*}

\subsection{Experimental Settings}

\textbf{Platform:}
We conduct the experiments on a server with 128 AMD EPYC 7543 32-Core Processors, 512GB main memory, and eight NVIDIA RTX A6000 GPUs, each with 48GB device memory. The operating system is Ubuntu 22.04. Our model is implemented in PyTorch 2.1.0.

\textbf{Evaluation Metrics:}
Following previous work \cite{gao2023out, qu2024connect, sarkar2024demystifying, wang2023galaxy, xu2023fourier, xu2023semantic}, we select metrics that best suit the characteristics of each dataset to ensure a comprehensive evaluation of model performance. We use F1 score and accuracy as evaluation metrics, depending on the nature of the task and dataset. The F1 score is used for iWildCam,  BirdCalls datasets to account for class imbalance, and we use accuracy for Camelyon17, Galaxy10, PACS, Digits-DG and OfficeHome.

\textbf{Method under Comparison:} 
We compare our method against various baseline approaches for domain generalization and out-of-domain (OOD) performance enhancement. These baselines include: 

\textbf{(1) Generic Augmentations and Domain Invariance Baselines.}
For \textbf{real-world datasets} iWildCam, Camelyon17 and Galaxy10, we compare our approach with several commonly used augmentation techniques: RandAugment \cite{cubuk2020randaugment}, CutMix \cite{yun2019cutmix}, MixUp \cite{zhang2017mixup},  Cutout \cite{devries2017improved}, FACT \cite{xu2023fourier}, and SAM \cite{xu2023semantic}. In the case of the BirdCalls dataset, we compare with MixUp, SpecAugment \cite{park2019specaugment}, random low/high pass filters, noise reduction \cite{sainburg2024noisereduce}, Jitter \cite{tellez2018whole,gao2023out}, FACT \cite{xu2023fourier}, and SAM \cite{xu2023semantic}. 
We also compare to LISA \cite{yao2022improving}, a method designed to encourage domain invariance by applying mixing augmentations to inputs of the same class across different domains. Additionally, we evaluate other domain invariance algorithms that do not rely on augmentation, including DANN \cite{long2018conditional, ganin2016domain}, DeepCORAL \cite{sun2016deep, sun2017correlation}, and ERM \cite{qu2024connect}, LP-FT \cite{kumar2022fine}. For the \textbf{benchmark datasets}, we compared with four SOTA baselines FACT \cite{xu2023fourier}, SAM \cite{xu2023semantic}, Su et al, \cite{su2024enhanced} (following the setting of DA + DeepCoral), and AGST \cite{wong2025approximate}. 

\textbf{(2) Dataset-Specific Targeted Augmentations.}
We also compare our method with dataset-specific targeted augmentations \cite{gao2023out} for real-world datasets: Copy-Paste for iWildCam, Stain Color Jitter for Camelyon17, and Copy-Paste + Jitter for BirdCalls. We also compare to Connect Later, which includes the dataset-specific targeted augmentations in finetuning after pretraining \cite{qu2024connect}. 

\textbf{Backbone Networks:} 
In section \ref{Comparison Results}, \ref{Ablation Study}, we use the same pretrained backbones as \cite{qu2024connect, gao2023out} for iWildCam (ResNet50 \cite{he2016deep}), Camelyon17 (DenseNet-121 \cite{huang2017densely}) and BirdCalls (EfficientNet-B0 \cite{tan2019efficientnet}). For Galaxy10, we follow \cite{wang2025galaxalign} to use ResNet18 \cite{he2016deep} for better performance. And we use the same ConvNet backbone as \cite{zhou2020deep, xu2023semantic, zhou2020learning, xu2023fourier} for Digits-DG, and ResNet18 \cite{he2016deep} for OfficeHome and PACS. We additionally evaluate our method in \ref{Generalization on Backbones} with ViT \cite{dosovitskiy2020image} and ConvNeXt \cite {liu2022convnet} backbone networks to verify that D-GAP generalizes beyond backbone choices.

\subsection{Comparison Results}
\label{Comparison Results}
\textbf{Results on Real-World Datasets.} Fig. \ref{points} summarizes the in-domain (ID) and out-of-domain (OOD) performance for all methods across four real-world datasets: iWildCam, Camelyon17, BirdCalls, and Galaxy10. Each point represents one method, colors and shapes represent method types, and our method is marked as a red star.
Our method consistently achieves the best OOD performance across all four real-world datasets, while maintaining a strong ID performance. Notably, our approach significantly outperforms all generic methods, including \textbf{generic augmentation} and \textbf{domain invariance} baselines that do not require dataset-specific design or prior knowledge. Compared to the best generic method (FACT) on iWildCam, our method improves OOD Macro F1 from 34.7\% to 36.8\%. On Camelyon17, our approach achieves 96.4\% OOD accuracy, outperforming the best generic augmentation (SAM at 92.2\%) and the best domain invariance method (LP-FT at 91.4\%). On BirdCall, OOD Macro F1 increases from 35.1\% (FACT) to 40.7\%, and on Galaxy10, our method improves OOD accuracy from 74.1\% (SAM) to 83.4\%. 

Moreover, our method also outperforms \textbf{dataset-specific augmentation} methods Connect Later + Copy-Paste on iWildCam, Connect Later + Stain Color Jitter on Camelyon17, and Connect Later + Copy-Paste + Jitter on BirdCalls, all of which require expert knowledge and manual design tailored for the dataset based on its characteristics. In contrast, our method is in a fully \textbf{dataset-agnostic} way, but still achieves superior or comparable results. These results show that our method is not only effective but also broadly applicable, without relying on dataset-specific analysis or prior knowledge.

\begin{table*}[h]
\centering
\footnotesize
\renewcommand{\arraystretch}{0.65}
\caption{Model accuracy of leave-one-domain-out evaluation on PACS and OfficeHome.}
\label{tab:pacs_officehome}
\setlength{\tabcolsep}{3.5pt}
\begin{tabular}{lcccc|c|cccc|c}
\toprule
\multirow{2}{*}{\textbf{Methods}}
& \multicolumn{5}{c|}{\textbf{PACS}}
& \multicolumn{5}{c}{\textbf{OfficeHome}} \\
\cmidrule(lr){2-6} \cmidrule(lr){7-11}
& \textbf{Art} & \textbf{Cartoon} & \textbf{Photo} & \textbf{Sketch} & \textbf{Avg.}
& \textbf{Art} & \textbf{Clipart} & \textbf{Product} & \textbf{Real} & \textbf{Avg.} \\
\midrule
DeepAll
& 84.94 & 76.98 & \underline{97.64} & 76.75 & 84.08
& 57.88 & 52.72 & 73.50 & 74.80 & 64.72 \\

FACT (2023) \cite{xu2023fourier}
& \textbf{89.68} & 81.06 & 97.02 & 83.76 & \underline{87.88}
& 60.34 & 54.85 & 74.48 & \underline{76.55} & 66.56 \\

SAM (2023) \cite{xu2023semantic}
& 85.25 & \underline{82.02} & 96.75 & \underline{85.04} & 87.27
& 60.79 & \textbf{55.47} & 74.37 & 76.37 & 66.75 \\

Su et al. (2024) \cite{su2024enhanced}
& 79.88 & 78.03 & 95.93 & 73.15 & 81.75
& \underline{64.13} & 54.83 & \underline{75.60} & 75.36 & \underline{67.71} \\

AGST (2025) \cite{wong2025approximate}
& 83.50 & 80.83 & 96.49 & 79.57 & 85.35
& 62.50 & 53.87 & 74.57 & 75.70 & 66.66 \\

\textbf{Ours}
& \underline{89.34} & \textbf{82.81} & \textbf{97.99} & \textbf{85.98} & \textbf{89.03}
& \textbf{64.82} & \underline{55.21} & \textbf{77.76} & \textbf{83.08} & \textbf{70.22} \\
\bottomrule
\end{tabular}
\end{table*}

\begin{table}[h]
\centering
\footnotesize
\renewcommand{\arraystretch}{0.5}
\caption{Model accuracy of leave-one-domain-out evaluation on Digits-DG.  }
\label{tab:digitsdg}
\setlength{\tabcolsep}{5pt}
\begin{tabular}{lcccc|c}
\toprule
\textbf{Methods} & \textbf{MNIST} & \textbf{MNIST-M} & \textbf{SVHN} & \textbf{SYN} & \textbf{Avg.} \\
\midrule
DeepAll        & 95.8        & 58.8       & 61.7        & 78.6        & 73.7 \\
FACT (2023) \cite{xu2023fourier}      & \underline{97.9} & 65.6 & 72.4 & 90.3 & 81.5 \\
SAM (2023) \cite{xu2023semantic}         & {97.8} & \underline{66.8} & {73.2} & {90.7} & {82.1} \\
Su et al. (2024) \cite{su2024enhanced} & 97.3 & 65.9 & \underline{74.5} & {92.5} & \underline{82.6} \\
AGST (2025) \cite{wong2025approximate} & 97.3 & 63.1 & 70.2 & \underline{93.2} & 81.0 \\
\textbf{Ours}  & \textbf{98.3} & \textbf{68.5} & \textbf{77.2} & \textbf{93.8} & \textbf{84.5}\\
\bottomrule
\end{tabular}
\end{table}


\begin{table*}[b]
\centering
\caption{
Ablation results across four real-world datasets. 
}
\footnotesize
\renewcommand{\arraystretch}{0.6}
\begin{tabular}{l c>{\columncolor{gray!15}[-5pt][-5pt]}c c>{\columncolor{gray!15}[-9pt][-9pt]}c c>{\columncolor{gray!15}[-3pt][-3pt]}c c>{\columncolor{gray!15}[-5pt][-5pt]}c c}
\toprule
& \multicolumn{2}{c}{{iWildCam (F1)}} & \multicolumn{2}{c}{{Camelyon17 (Acc)}} & \multicolumn{2}{c}{{BirdCall (F1)}} & \multicolumn{2}{c}{{Galaxy10 (Acc)}} \\
\cmidrule(lr){2-3} \cmidrule(lr){4-5} \cmidrule(lr){6-7} \cmidrule(lr){8-9}
& {ID} & {OOD} & {ID} & {OOD} & {ID} & {OOD} & {ID} & {OOD} \\
\midrule
Ours w/o Augmentations & 46.4 & 31.2 & 92.3 & 91.4 & 73.1 & 30.1 & 93.6 & 54.3  \\
Pixel-only & 45.4 & 29.1 & 87.5 & 69.8 & 75.2 & 32.4 &95.5 & 68.2 \\
Frequency-only & \textbf{52.0} & 35.6 & \underline{98.3} & 95.7 & 73.4 & 37.6 & 95.1 & 77.9 \\
Mask-low  & 51.2 & \underline{36.0} & \underline{98.3} & \underline{96.1} & \textbf{75.5} & \underline{38.4} & 95.1 & \underline{80.5}  \\
Mask-high & 51.5 & 35.4 & 97.9 & 95.4 & 75.1 & 36.4 & 95.0 & 79.5 \\
Mask-ring & \underline{51.9} & 34.3 & 98.2 & 94.0 & 74.9 & 36.5 & \textbf{95.7} & 77.1 \\
Ours & \underline{51.9} & \textbf{36.8} & \textbf{98.6} & \textbf{96.4} & \underline{75.4} & \textbf{40.7} & \underline{95.3} & \textbf{83.4} \\
\bottomrule
\end{tabular}
\label{tab:ablation}
\end{table*}

\textbf{Results on Common Benchmarks.}
Tables~\ref{tab:pacs_officehome} and \ref{tab:digitsdg} summarize the leave-one-domain-out
results on PACS, Office-Home and Digits-DG (Mean). The results of FACT and SAM in these tables are from the source paper \cite{xu2023semantic} for fair comparison, as our re-implementations under identical settings produced slightly lower numbers. DeepAll refers to vanilla backbones trained from source data \cite{xu2021fourier}. In Tables~\ref{tab:pacs_officehome} and \ref{tab:digitsdg}, our method achieves the highest average accuracy on all three benchmarks. These results show that the proposed dual-space targeted augmentation is dataset-agnostic and robust to different domain variations, while our focus remains on real-world deployment or environment shifts. The results with standard deviation are listed in Appendix Sec. \ref{Additional Comparison Results}.

\subsection{Ablation Study}
\label{Ablation Study}
Table~\ref{tab:ablation} presents ablation results of the effects of pixel-space and frequency-space mixing and the impact of the gradient-guided mechanism.
Across the four real-world datasets, our full method achieves the highest OOD performance, outperforming Ours w/o Augmentations and all single-modality variants. We also replace the gradient-guided map with fixed unguided maps with frequency masks: low-frequency mask, high-frequency mask, and ring mask. 
Frequency-only mixing consistently outperforms pixel-only mixing, highlighting the importance of frequency-space perturbation. Replacing the gradient-guided mechanism with maps with low/high/ring masks leads to a consistent drop in OOD performance, especially on BirdCalls and Galaxy10, confirming that this adaptive, sensitivity-aware guidance is essential for robust domain adaptation.


\subsection{Generalization on Backbone Networks}
\label{Generalization on Backbones}
\begin{figure}[h]
\centerline{\includegraphics[height=3.8cm]{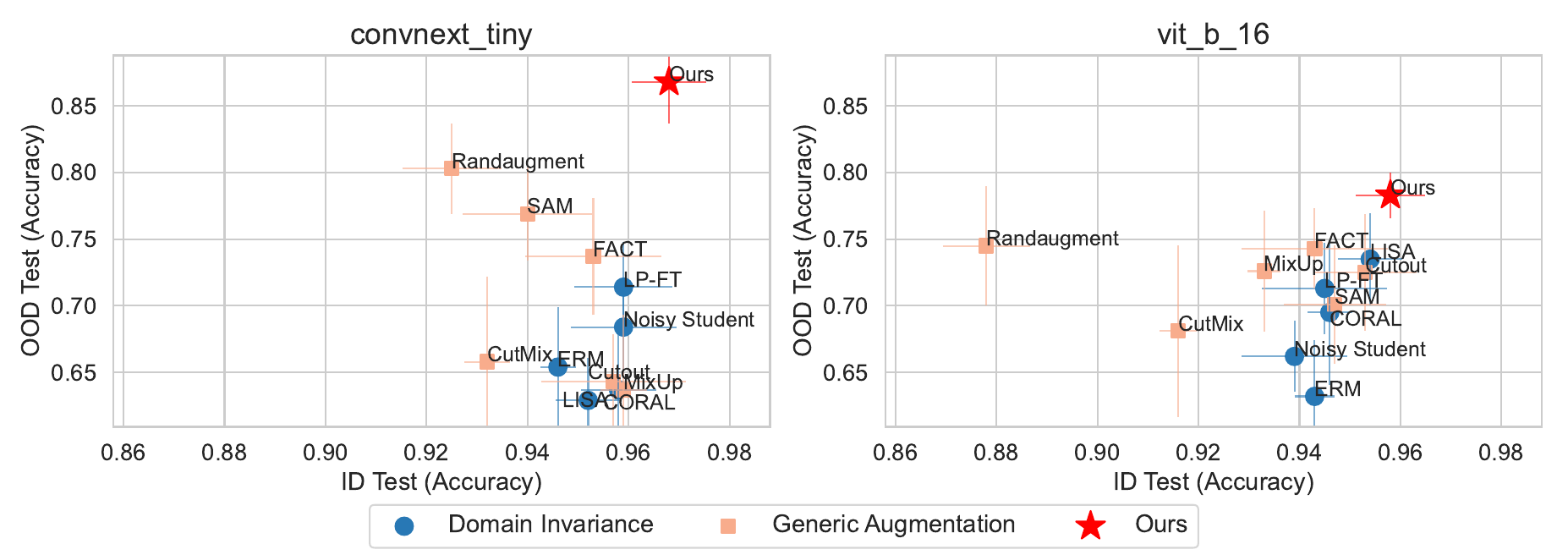}}
\caption{Performance of ConvNeXt and ViT on Galaxy10.}
\label{idood_ga10}
\end{figure}

To verify that D-GAP generalizes beyond backbone networks, we additionally evaluate with ConvNeXt-Tiny \cite{liu2022convnet} and ViT-B/16 \cite{dosovitskiy2020image}.
Table~\ref{tab:backbone_results} shows that D-GAP consistently improves OOD performance on PACS, Digits-DG, and OfficeHome under both backbones.
Fig.~\ref{idood_ga10} further visualizes the ID--OOD trade-off on Galaxy10, where D-GAP remains SOTA performance on OOD robustness. 

\vspace{-5pt}

\begin{table}[h]
\centering
\renewcommand{\arraystretch}{0.55}
\caption{Average OOD performance on PACS, Digits-DG, and OfficeHome with ConvNeXt and ViT.}
\footnotesize
\begin{tabular}{lcccccc}
\toprule
\multirow{2}{*}{Method} & \multicolumn{2}{c}{PACS} & \multicolumn{2}{c}{Digits} & \multicolumn{2}{c}{OfficeHome} \\
\cmidrule(lr){2-3}\cmidrule(lr){4-5}\cmidrule(lr){6-7}
 & ConvNeXt & ViT & ConvNeXt & ViT & ConvNeXt & ViT \\
\midrule
SAM (2023) \cite{xu2023semantic} & 85.3 & 85.4 & 88.0 & 85.1 & 73.7 & 63.8 \\
Su et al. (2024) \cite{su2024enhanced} & 87.9 & 75.3 & 88.9 & 85.6 & 73.9 & 64.6 \\
AGST (2025) \cite{wong2025approximate} & 87.4 & 79.0 & 88.5 & 85.8 & 73.7 & 64.4 \\
Ours & \textbf{88.7} & \textbf{86.4} & \textbf{89.8} & \textbf{86.8} & \textbf{75.0} & \textbf{66.0} \\
\bottomrule
\end{tabular}
\label{tab:backbone_results}
\vspace{-3pt}
\end{table}


\begin{table}[t]
\centering
\caption{Jaccard similarity of top-ranked high-gradient frequency regions.
SC-CD denotes same-class cross-domain pairs, while DC-SD denotes
different-class same-domain pairs.}
\label{tab:gradient_domain_analysis}
\footnotesize
\begin{tabular}{lcccc}
\toprule
& \multicolumn{2}{c}{Galaxy10} 
& \multicolumn{2}{c}{OfficeHome} \\
\cmidrule(lr){2-3}
\cmidrule(lr){4-5}
High-gradient region 
& SC-CD & DC-SD 
& SC-CD & DC-SD \\
\midrule
Top 10\% 
& 0.5449 & \textbf{0.5551}
& 0.2774 & \textbf{0.3035} \\

Top 20\% 
& 0.5485 & \textbf{0.5734}
& 0.3332 & \textbf{0.3554} \\

Top 30\% 
& 0.5497 & \textbf{0.5847}
& 0.4372 & \textbf{0.4558} \\
\bottomrule
\end{tabular}
\end{table}

\subsection{Analysis of Gradient-based Sensitivity}
\label{sec:gradient_analysis}

A high gradient magnitude measures local model sensitivity rather than directly
identifying domain-specific or spurious components. We therefore compare the
Jaccard overlap of high-gradient regions between same-class cross-domain
(SC-CD) and different-class same-domain (DC-SD) pairs. As shown in
Table~\ref{tab:gradient_domain_analysis}, DC-SD pairs consistently exhibit
higher similarity across both datasets and all thresholds, suggesting that
high-gradient frequency locations contain domain-dependent information rather
than being determined solely by class identity. This provides empirical support
for using model sensitivity to guide cross-domain spectral perturbation.

\subsection{Empirical Evaluations of Connectivity}
\label{Empirical Evaluations of Connectivity}
To better understand why our method is effective on improving OOD robustness, we empirically evaluate the connectivity measures (details are in Appendix Sec. \ref{appendix_connectivity}). 
We present connectivity ratios $\alpha/\gamma$ and $\beta/\gamma$ on datasets Camelyon17 and iWildCam in Table \ref{connect}. 
Our method achieves the highest $\alpha/\gamma$ value on both datasets, showing more consistent semantic alignment across domains as well as more effective randomization of spurious domain-dependent features $x_{d\text{:spu}}$. However, a moderate $\beta/\gamma$ indicates our method may also slightly perturbs class-relevant features $x_{\text{obj}}, x_{d\text{:robust}}$. However, from Table \ref{tab:ablation} and Fig.\ref{points}, our method still achieves the best OOD performance.
The result is consistent with the estimation proposed by Shen et al.~\cite{shen2022connect}, where the target accuracy is estimated to be $ (\alpha/\gamma)^{w_1} \cdot (\beta/\gamma)^{w_2},
$
with higher weight on $\alpha/\gamma$ ($w_1 > w_2$).
These results show that D-GAP achieves a balance of preservation of class-relevant features and perturbation of domain-relevant features.



\begin{table*}[h]
\centering
\footnotesize
\renewcommand{\arraystretch}{0.6}
\caption{Connectivity values across class, domain, and both on Camelyon17 and iWildCam datasets.}
\label{connect}
\setlength{\tabcolsep}{4pt}
\begin{tabular}{l|ccc|cc|l|ccc|cc}
\toprule
\multicolumn{6}{c|}{\textbf{Camelyon17}} 
& \multicolumn{6}{c}{\textbf{iWildCam}} \\
\cmidrule(lr){1-6} \cmidrule(lr){7-12}
\textbf{Augmentations} 
& $\alpha$ & $\beta$ & $\gamma$ & $\alpha/\gamma$ & $\beta/\gamma$
& \textbf{Augmentations}
& $\alpha$ & $\beta$ & $\gamma$ & $\alpha/\gamma$ & $\beta/\gamma$ \\
\midrule

w/o Augmentations 
& 0.015 & 0.189 & 0.002 & 7.50 & 94.5
& w/o Augmentations
& 0.007 & 0.008 & 0.021 & 0.33 & 0.38 \\

RandAugment
& 0.181 & 0.186 & 0.114 & 1.59 & 1.63
& RandAugment
& 0.224 & 0.125 & 0.130 & 1.72 & 0.96 \\

Stain Color Jitter
& 0.029 & 0.116 & 0.004 & 7.25 & 29.0
& Copy-Paste
& 0.254 & 0.031 & 0.063 & 4.03 & 0.49 \\

\textbf{Ours}
& 0.120 & 0.073 & 0.003 & 40.0 & 24.3
& \textbf{Ours}
& 0.212 & 0.030 & 0.051 & 4.16 & 0.59 \\

\bottomrule
\end{tabular}
\end{table*}

\paragraph{Computational cost.}
Table~\ref{tab:compute} compares the computational overhead under a unified
experimental setting. The additional cost of D-GAP mainly comes from computing task gradients with respect to the Fourier amplitude in order to construct the model-response-adaptive sensitivity map.

\begin{table}[h]
\centering
\caption{Computational cost comparison under the same setting. }
\label{tab:compute}
\footnotesize
\setlength{\tabcolsep}{4pt}
\begin{tabular}{lccc}
\toprule
Method &
Time / Iter. (ms) &
Throughput (img/s) &
Peak GPU Mem. (GB) \\
\midrule
ERM   & $24.46 \pm 0.17$ & $1308.3 \pm 9.1$ & 0.86 \\
MixUp & $24.71 \pm 0.16$ & $1294.9 \pm 8.1$ & 0.88 \\
DANN  & $48.62 \pm 0.24$ & $658.2 \pm 3.2$  & 1.55 \\
LISA  & $66.28 \pm 0.19$ & $482.8 \pm 1.3$  & 0.88 \\
SAM   & $48.00 \pm 0.30$ & $666.6 \pm 4.2$  & 0.91 \\
FACT  & $63.00 \pm 0.31$ & $507.9 \pm 2.5$  & 1.94 \\
D-GAP (Ours)
      & $90.46 \pm 2.38$ & $353.9 \pm 9.2$  & 0.86 \\
\bottomrule
\end{tabular}
\end{table}

\vspace{-3pt}
\section{Conclusion, Limitations and Future Work} 
\vspace{-3pt}
We propose D-GAP, providing a simple and adaptive approach for OOD robustness in visual domain adaptation without dataset-specific analysis. By operating in both frequency and pixel spaces and adaptively interpolating amplitude values, our method achieves notable improvements on multiple real-world datasets and domain adaptation benchmark datasets. One limitation is that the current framework needs additional gradient computation. However, this cost comes from the model-response-adaptive nature of the method, which is also the key distinction from fixed or random Fourier mixing. 
Future work may reduce this overhead by updating sensitivity maps periodically, using low-resolution spectral proxies, or learning lightweight sensitivity predictors. 


\section*{Acknowledgments}
This work was supported by the National Nature Science Foundation of China (NSFC) No. 62372393.

\bibliography{ref}

\newpage

\appendix
\begin{center}
    \Large{\textbf{Appendix}}
\end{center}

\section{Additional Comparison Results}
\label{Additional Comparison Results}
Due to page limitations, we present the results in Tables \ref{tab:pacs_officehome} and \ref{tab:digitsdg} with the additional variance value over five random seeds here in Tables \ref{tab:pacs_var}, \ref{tab:digitsdg_var}, and \ref{tab:officehome_var}. Results of DeepAll, FACT and SAM in these tables are copied from the source paper \cite{xu2023semantic} for fair comparison, as our re-implementations under identical settings produced slightly lower numbers. As a result, we did not provide standard deviation values for DeepAll, FACT and SAM here for fair comparison.

\begin{table}[h]
\centering
\footnotesize
\renewcommand{\arraystretch}{0.9}
\caption{Model accuracy of leave-one-domain-out evaluation on Digits-DG (mean and standard deviation).  }
\label{tab:digitsdg_var}
\setlength{\tabcolsep}{5pt}
\begin{tabular}{lcccc|c}
\toprule
\textbf{Methods} & \textbf{MNIST} & \textbf{MNIST-M} & \textbf{SVHN} & \textbf{SYN} & \textbf{Avg.} \\
\midrule
DeepAll        & 95.8        & 58.8       & 61.7        & 78.6        & 73.7 \\
FACT (2023) \cite{xu2023fourier}      & \underline{97.9} & 65.6 & 72.4 & 90.3 & 81.5 \\
SAM (2023) \cite{xu2023semantic}         & {97.8} & \underline{66.8} & {73.2} & {90.7} & {82.1} \\
Su et al. (2024) \cite{su2024enhanced} & 97.3 $\pm$ 0.33 & 65.9 $\pm$ 0.51 & \underline{74.5 $\pm$ 0.48} & {92.5 $\pm$ 0.31} & \underline{82.6} \\
AGST (2025) \cite{wong2025approximate} & 97.3 $\pm$ 0.14 & 63.1  $\pm$ 0.45 & 70.2  $\pm$ 0.78 & \underline{93.2 $\pm$ 0.37} & 81.0 \\
\textbf{Ours}  & \textbf{98.3 $\pm$ 0.12} & \textbf{68.5  $\pm$ 0.26} & \textbf{77.2 $\pm$ 0.53} & \textbf{93.8 $\pm$ 0.24} & \textbf{84.5}\\
\bottomrule
\end{tabular}
\end{table}

\begin{table}[h]
\centering
\footnotesize
\renewcommand{\arraystretch}{0.9}
\caption{Model accuracy of leave-one-domain-out evaluation on OfficeHome (mean and std). }
\label{tab:officehome_var}
\setlength{\tabcolsep}{6pt}
\begin{tabular}{lcccc|c}
\toprule
Methods & Art & Clipart & Product & Real & Avg. \\
\midrule
DeepAll        & 57.88 & 52.72 & 73.50 & 74.80 & 64.72 \\
FACT (2023) \cite{xu2023fourier}    & 60.34 & 54.85 & {74.48} & \underline{76.55} & 66.56 \\
SAM (2023) \cite{xu2023semantic} & {60.79} & \textbf{55.47} & 74.37 & 76.37 & {66.75} \\
Su et al. (2024) \cite{su2024enhanced} & \underline{64.13 $\pm$ 0.50} & 54.83 $\pm$ 0.49 & \underline{75.60 $\pm$ 0.61} & 75.36 $\pm$ 1.04 & \underline{67.71} \\
AGST (2025) \cite{wong2025approximate} & 62.50 $\pm$ 0.61 & 53.87 $\pm$ 0.29 & 74.57 $\pm$ 0.48 & 75.70 $\pm$ 0.78 & 66.66 \\
\textbf{Ours} & \textbf{64.82 $\pm$ 0.55} & \underline{55.21 $\pm$ 0.37} & \textbf{77.76 $\pm$ 0.43} & \textbf{83.08 $\pm$ 0.25} & \textbf{70.22}\\
\bottomrule
\end{tabular}
\end{table}

\begin{table}[h]
\centering
\footnotesize
\renewcommand{\arraystretch}{0.9}
\caption{Model accuracy of leave-one-domain-out evaluation on PACS (mean and std). }
\label{tab:pacs_var}
\setlength{\tabcolsep}{5pt}
\begin{tabular}{lcccc|c}
\toprule
\textbf{Methods} & \textbf{Art} & \textbf{Cartoon} & \textbf{Photo} & \textbf{Sketch} & \textbf{Avg.} \\
\midrule
DeepAll       & 84.94                     & 76.98                     & \underline{97.64}          & 76.75                     & 84.08 \\
FACT (2023) \cite{xu2023fourier}    & \textbf{89.68}            & 81.06                     & 97.02                     & {83.76}         & \underline{87.88} \\
SAM (2023) \cite{xu2023semantic} & 85.25 & \underline{82.02} & 96.75 & \underline{85.04} & 87.27 \\
Su et al. (2024) \cite{su2024enhanced} & 79.88 $\pm$ 0.53  & 78.03 $\pm$ 0.57 & 95.93 $\pm$ 0.21 & 73.15 $\pm$ 0.39 & 81.75 \\
AGST (2025) \cite{wong2025approximate} & 83.50 $\pm$ 0.76 & 80.83  $\pm$ 0.48 & 96.49 $\pm$ 0.23 & 79.57 $\pm$ 0.62 & 85.35 \\
\textbf{Ours} & \underline{89.34 $\pm$ 0.34} & \textbf{82.81 $\pm$ 0.63} & \textbf{97.99 $\pm$ 0.14} & \textbf{85.98 $\pm$ 0.19} & \textbf{89.03} \\ 
\bottomrule
\end{tabular}
\end{table}

\begin{figure*}[h]
    \centering
    \begin{subfigure}{}
        \centering
        \includegraphics[width=0.45\textwidth]{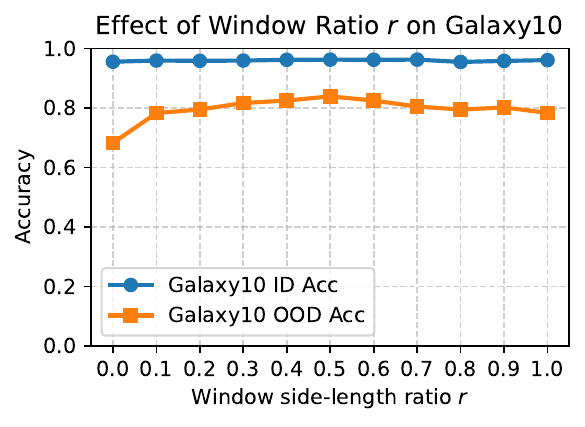}
    \end{subfigure}
    \hfill
    \begin{subfigure}{}
        \centering
        \includegraphics[width=0.44\textwidth]{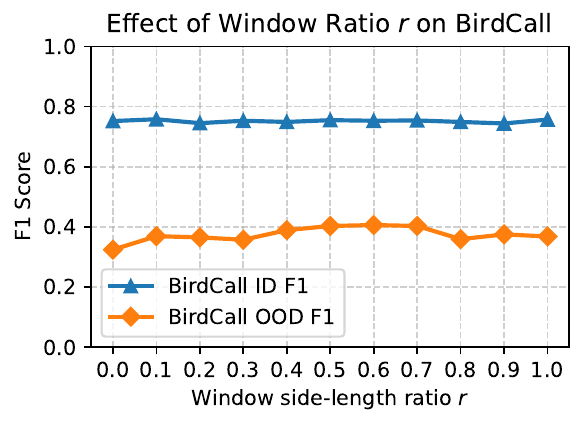}
    \end{subfigure}
    \caption{Effect of the frequency-window side-length ratio $r$ on Galaxy10 and BirdCalls datasets.}
    \label{fig:omega_ablation}
\end{figure*}

\section{Analysis over Hyperparameters}
\label{Analysis over Hyperparameters}
\textbf{Effect of the frequency-window ratio.} We further investigate the influence of the frequency-window side-length ratio $r$, which controls the size of the square region $\Omega_r$ used for amplitude mixing in the Fourier space (in Equation \ref{eq3}). As shown in Fig.~\ref{fig:omega_ablation}, the choice of $r$ has only a minor effect on the in-domain (ID) performance. On both Galaxy10 and BirdCall, the ID accuracy/F1 remains relatively stable across different values of $r$. For out-of-domain (OOD) generalization, the performance is more sensitive to $r$, but the variation remains moderate as long as the frequency window is not zero. In particular, both datasets achieve strong OOD performance when $r$ is in the range of $0.4$ to $0.6$. Although other none-zero choices of $r$ are slightly inferior to this range, they still consistently outperform competing baseline methods, demonstrating that the proposed augmentation is robust to the exact selection of the window size. In the experiments in our paper, we uniformly set $r=0.5$ for all datasets, with which our method achieves state-of-the-art results. And we recommend using $r=0.5$ as the default setting in further practical applications.

\begin{figure*}[h]
\centerline{\includegraphics[height=9cm]{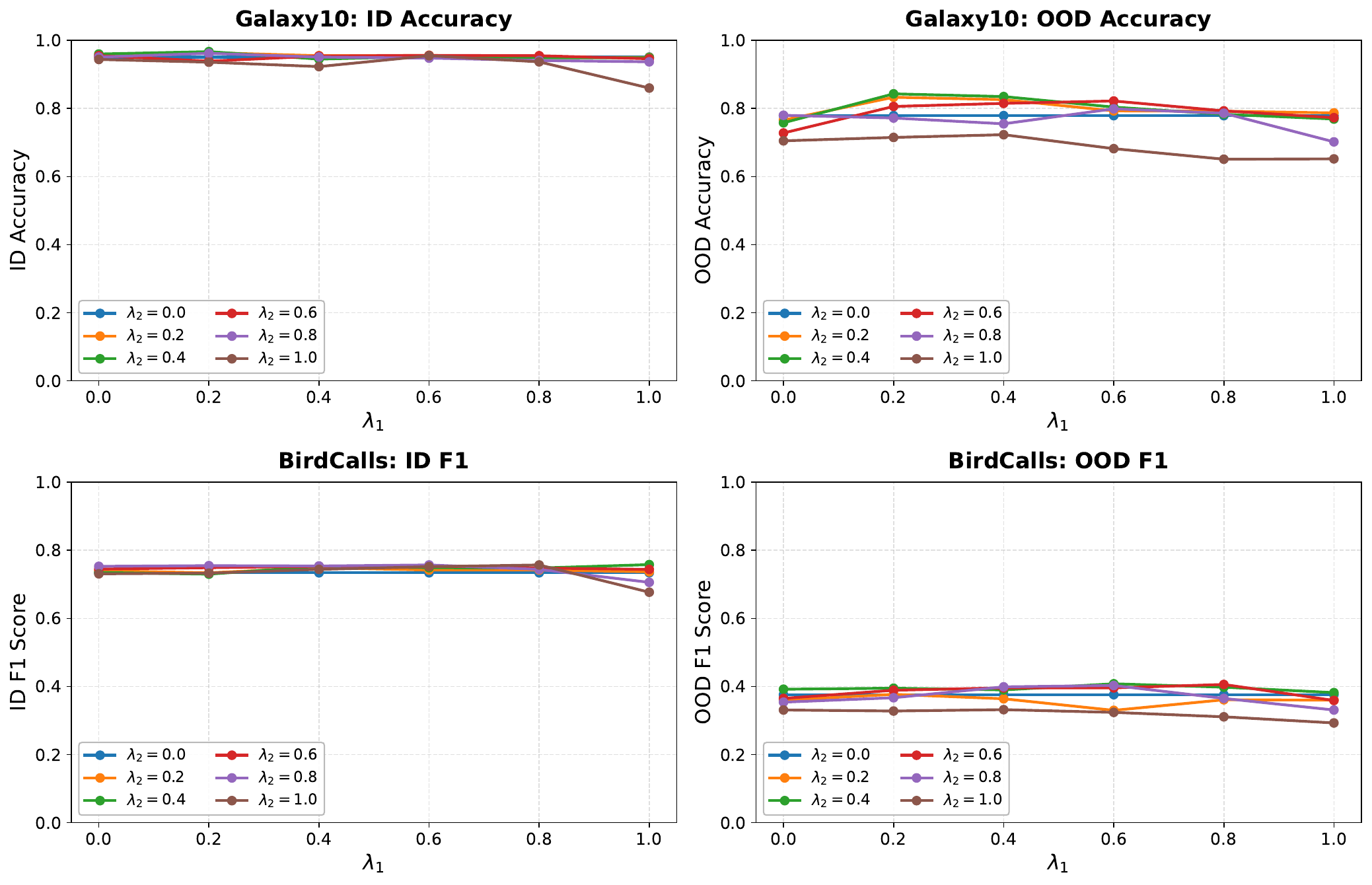}}
\caption{Effect of the mixing ratios $\lambda_1$ and $\lambda_2$ on Galaxy10 and BirdCalls datasets.}
\label{fig:lambda_ablation}
\end{figure*} 

\textbf{Effect of the mixing ratio $\lambda_1$ and $\lambda_2$.} 
We analyze the sensitivity of the proposed method to the mixing ratios $\lambda_1$ and $\lambda_2$ by varying both values from $0$ to $1$. As shown in Fig.~\ref{fig:lambda_ablation}, the ID performance remains stable across different parameter pairs, suggesting that the proposed augmentation does not degrade source-domain learning. The OOD performance is slightly more sensitive, but remains consistently strong as long as $\lambda_1$ and $\lambda_2$ are not set to extreme values such as $0$ or $1$. Most non-extreme combinations achieve competitive results and still outperform the baseline methods, indicating that the method is robust to the exact choice of these two hyperparameters. Overall, we recommend setting $\lambda_1 \in [0.2,0.6]$ and $\lambda_2 \in [0.2,0.6]$, while avoiding overly weak or overly aggressive mixing.

\section{Related Work}
\label{Related Work}
\subsection{Pretraining for OOD Robustness} 
Pretraining on large-scale unlabeled data has emerged as a promising OOD generalization method \cite{caron2020unsupervised, shen2022connect, radford2021learning, sagawaextending}. Contrastive pretraining, which encourages feature alignment across domains using generic augmentations, outperforms traditional domain adaptation methods that enforce explicit domain invariance \cite{kang2019contrastive, hoffman2018cycada}. However, the standard fine-tuning with generic augmentations often leads to unsatisfying OOD performance due to misalignment with specific distribution shifts \cite{qu2024connect, kumar2022fine}. Although Connect Later \cite{qu2024connect} improves OOD robustness by applying targeted augmentations during linear probing then fine-tuning (LP-FT) \cite{kumar2022fine}, it still relies heavily on manually designed augmentations and cannot easily generalize without shift-specific analysis. 

\subsection{Augmentations}
For a given input $x \in \mathcal{X}$, augmented variants $x’ \sim \mathcal{A}(\cdot \mid x)$ are often used to enhance robustness \cite{hendrycksaugmix, hendrycks2021many}.
Data augmentation has been widely used to improve model robustness to label-invariant transformations such as translation or rotation \cite{hendrycksaugmix, rebuffi2021data, ng2020ssmba}. Traditional augmentation strategies typically apply generic perturbations to increase input diversity \cite{simard2003best, krizhevsky2012imagenet, cubuk2020randaugment, devries2017improved, zhang2017mixup}. However, recent studies highlight that the choice of augmentation type significantly impacts model generalization \cite{chen2020simple, xie2020unsuperviseda}.

In semi-supervised learning, augmentations have also been utilized in the self-training paradigm to improve generalization by training on pseudo-labeled data \cite{xie2020self, sohn2020fixmatch, yang2021interactive}. Nevertheless, Qu et al. \cite{qu2024connect} show that a self-training baseline using pseudo-labels generated from an ERM model with targeted augmentations still underperforms compared to Connect Later \cite{qu2024connect}. This underscores the importance of pretrained representations in achieving robustness gains. Connect Later \cite{qu2024connect} leverages targeted augmentations not only as a data transformation tool, but as a design interface to incorporate knowledge of the distribution shift while still benefiting from pretrained features.

Recent work in domain shift problems emphasizes the effectiveness of \textit{targeted augmentations}—those informed by the nature of the distribution shift—over generic or even target-aware augmentations such as CutMix \cite{gao2023out, yun2019cutmix}. While \cite{gao2023out} focuses on domain generalization where the target domain is unknown, some recent work \cite{qu2024connect,gao2023out} studies the domain adaptation setting with access to unlabeled target data. The targeted augmentations are effective but rely on manually designing and are dataset-specialized.

Frequency-domain augmentation has been widely used to improve robustness under domain shifts. 
FDA \cite{yang2020fda49} replaces the low-frequency amplitude of source images with target-domain spectra, and \cite{yang2020phase48} preserves phase for semantic consistency, but both rely on manually constrained spectral rules. 
FSDR \cite{huang2021fsdr} randomizes domain-variant frequency components, but its perturbation is based on domain-level statistics rather than the model's local task sensitivity. 
HybridAugment++ \cite{yucel2023hybridaugment} unifies multiple spectrum perturbations, FSDA \cite{nam2024fsda} learns frequency re-scaling from spectral patterns, and Vital Phase Augmentation \cite{lee2024domain} augments phase variations; however, these methods do not explicitly inject target-domain information according to frequency-wise model responses. FACT \cite{xu2023fourier} mixes Fourier amplitudes to encourage phase-preserved semantic learning, while SAM \cite{xu2023semantic} further introduces semantic-aware amplitude/phase mixup; however, their mixing strategies are still predefined rather than model-response-adaptive. 
In conclusion, existing methods \cite{xu2023semantic, xu2023fourier, yang2020fda49, yang2020phase48, yucel2023hybridaugment, nam2024fsda, lee2024domain, salehnia2025frequency, huang2021fsdr} typically use fixed, random, manually constrained, or data-statistic-driven perturbations, limiting targeted domain intervention. 
In contrast, D-GAP computes task-gradient-based amplitude sensitivity maps to adaptively control frequency-wise target-domain interpolation, and further combines pixel-space blending to compensate local spatial details, enabling more targeted and flexible OOD augmentation.

\subsection{On the Importance of Phase Information}
A series of early works \cite{oppenheim1979phase33, oppenheim1981importance34, piotrowski1982demonstration36,hansen2007structural12} have highlighted that in the Fourier spectrum of visual signals, the phase component is critical for preserving high-level semantics, while the amplitude primarily captures low-level appearance statistics. Recent advances further explore this perspective in domain adaptation. For example, \cite{yang2020fda49} proposes to replace a local region in the centralized amplitude spectrum of a source image with that of a target image to synthesize target-style images for training. In parallel, \cite{yang2020phase48} introduces phase consistency as a constraint during source-target translation, demonstrating its superiority over traditional cycle consistency \cite{hoffman2018cycada16} in domain adaptation tasks, especially for semantic segmentation. Xu et al. \cite{xu2023fourier} incorporate a Fourier-based strategy that emphasizes phase preservation to enhance generalization performance under domain shift. Inspired by this, we propose to use phase in frequency space to preserve the task-relevant features in our augmentations.

\section{Dataset Details}
\label{Dataset Details}
\paragraph{Species Classification from Camera Trap Images (iWildCam).} 
In the iWildCam dataset \cite{beery2021iwildcam, koh2021wilds}, the task is to classify an animal species \(y\) from images \(x\) captured by static camera traps. The dataset consists of 243 camera locations in \(D^{train}\). Although images from the same camera share similar backgrounds, domain-dependent features \(x_{d: robust}\), such as habitat, vary across domains. For example, cameras in Kenyan savannas (cameras 23 and 97) and a Guatemalan forest (camera 54) have differing background features. This requires the model to focus on foreground features (e.g., animal species) for accurate prediction, as background habitat often doesn't provide useful information for classification \cite{gao2023out, qu2024connect}.

\paragraph{Tumor Identification in Histopathology Slides (Camelyon17).}
In Camelyon17 \cite{beery2021iwildcam, koh2021wilds}, the task is to classify tumor presence in histopathology slides. These slides come from three hospitals, each with varying imaging protocols and staining techniques that lead to domain-specific color variations. For example, some hospitals have mostly early-stage pN1 breast cancer, while others have later-stage pN2 cases. These domain-specific features, such as tumor size and density, correlate with the cancer stage \(y\), requiring the model to account for these domain shifts in order to generalize effectively \cite{gao2023out, qu2024connect}.

\paragraph{Bird Species Recognition from Audio Recordings (BirdCalls).}
The BirdCalls dataset \cite{joly2022overview, gao2023out} consists of audio recordings from nine microphones located in different regions. The task is to classify the bird species \(y\) vocalizing in each audio clip \(x\). While the bird species information is crucial, low-level noise and microphone settings (e.g., gain levels) often introduce irrelevant domain-specific features. These features, such as background noises from different habitats, should be ignored by the model, which should focus on bird calls for accurate classification. Randomizing microphone-related features ensures that the model generalizes across diverse recording environments \cite{gao2023out}.

\paragraph{Astronomical Morphology Classification (Galaxy10 DECaLS\&SDSS).}
Galaxy 10 DECaLS and SDSS are two galaxy morphology classification datasets containing images from two different surveys \cite{Galaxy10}. They differ in image quality, resolution, and color distribution due to differences in telescope instrumentation. We treat Galaxy10-DECaLS as the source domain and Galaxy10-SDSS as the target domain, and evaluate cross-domain generalization without using any target-domain labels during training.

\paragraph{Additional Benchmark Datasets.}
In addition to the four real-world datasets used throughout this paper, we further evaluate our method
on three commonly used domain adaptation/generalization benchmarks: \textbf{PACS}~\cite{li2017deeper},
\textbf{Office-Home}~\cite{venkateswara2017deep}, and \textbf{Digits-DG}~\cite{zhou2020deep}. Compared to our four \emph{real-world} datasets that cover
diverse modalities and deployment-related shifts (wildlife, histopathology, audio, and astronomical imaging),
these common benchmarks focus on visual recognition with well-studied style variations
(PACS: \{Art, Cartoon, Photo, Sketch\}; Office-Home: \{Art, Clipart, Product, Real-World\};
Digits-DG: multiple digits domains including
MNIST~\cite{lecun2002gradient}, MNIST-M~\cite{ganin2015unsupervised}, SVHN~\cite{netzer2011reading}, and
SYN~\cite{ganin2015unsupervised}).

\begin{table}[h]
\centering
\small
\caption{Hyperparameter settings for baseline methods on Galaxy10.}
\renewcommand{\arraystretch}{0.9}
\label{settings}
\begin{tabular}{ll}
\toprule
\textbf{Method} & \textbf{Hyperparameters} \\
\midrule
ERM & \begin{tabular}[t]{@{}l@{}}
Learning rate $= 10^{-4}$ \\
Weight decay $= 0$
\end{tabular} \\
\midrule
LISA & \begin{tabular}[t]{@{}l@{}}
Learning rate $= 10^{-4}$ \\
Weight decay $= 0$ \\
Transform probability $= 0.9$
\end{tabular} \\
\midrule
MixUp & \begin{tabular}[t]{@{}l@{}}
Learning rate $= 10^{-4}$ \\
Weight decay $= 0$ \\
Transform probability $= 0.9$ \\
$\alpha = 0.4$
\end{tabular} \\
\midrule
CutMix & \begin{tabular}[t]{@{}l@{}}
Learning rate $= 10^{-4}$ \\
Weight decay $= 0$ \\
Transform probability $= 0.9$ \\
$\alpha = 0.4$
\end{tabular} \\
\midrule
Cutout & \begin{tabular}[t]{@{}l@{}}
Learning rate $= 10^{-4}$ \\
Weight decay $= 0$ \\
Transform probability $= 0.9$ \\
$\alpha = 0.4$
\end{tabular} \\
\midrule
Randaugment & \begin{tabular}[t]{@{}l@{}}
Learning rate $= 10^{-4}$ \\
Transform probability $= 0.9$ \\
Version = $Original$
\end{tabular} \\
\midrule
CORAL & \begin{tabular}[t]{@{}l@{}}
Learning rate $= 10^{-4}$ \\
Coral weight $= 1.0$ 
\end{tabular} \\
\midrule
Noisy Student & \begin{tabular}[t]{@{}l@{}}
Learning rate $= 10^{-4}$ \\
soft label $=False$
\end{tabular} \\
\midrule
DANN & \begin{tabular}[t]{@{}l@{}}
Classifier learning rate $= 10^{-4}$ \\
Discriminator learning rate $= 10^{-4}$ \\
GRL$\lambda = 2 / (1 + exp(-10·epoch/15)) - 1$
\end{tabular} \\
\bottomrule
\end{tabular}
\end{table}

\section{Experimental Details}
\label{Experimental Details}

\subsection{D-GAP}
We listed some important implementation details of our method as follows. Other implementation details can be find in our code in the supplementary material. And we will release the code after this paper is published.

\paragraph{iWildCam.}
We use ResNet-50 \cite{he2016deep} pretrained with SwAV \cite{caron2020unsupervised} as the backbone \cite{qu2024connect} in Section 4.2 and 4.3. For pretrained models, we follow LP-FT \cite{kumar2022fine, qu2024connect}: a linear probe is trained for 10 epochs using Adam, and its weights are used to initialize the full model, which is then fine-tuned for another 20 epochs. Following \cite{qu2024connect}, we perform 10 hyperparameter search trials by sampling: linear probe learning rate $10^{\text{Uniform}[-3,-2]}$, fine-tuning learning rate $10^{\text{Uniform}[-5,-2]}$, and augmentation probability from $\text{Uniform}[0.5, 0.9]$. The configuration with the best OOD validation performance is selected for testing. To mitigate outlier variance, our method is averaged over 5 seeds. Other hyperparameters are the same as those of Qu et al. \cite{qu2024connect}.

\paragraph{Camelyon17.}
We conduct experiments on the unlabeled Camelyon17 dataset \cite{sagawaextending} using DenseNet121 \cite{huang2017densely} pretrained with SwAV \cite{caron2020unsupervised} in Sections 4.2 and 4.3, following the protocol in \cite{qu2024connect, gao2023out}. For pretrained models, we first apply 10 epochs of linear probing before full fine-tuning for 20 epochs, and select the linear probe checkpoint with the highest OOD validation accuracy for initialization. For hyperparameter tuning, following \cite{qu2024connect}, the following hyperparameters are sampled: linear probe learning rate from $10^{\text{Uniform}[-3,-2]}$, fine-tuning learning rate from $10^{\text{Uniform}[-5,-2]}$, augmentation probability from $\text{Uniform}[0.5, 0.9]$, and augmentation strength from $\text{Uniform}[0.05, 0.1]$. The best hyperparameter configuration is selected based on OOD validation performance. All results are averaged over 5 random seeds.  Other hyperparameters are the same as those of Qu et al. \cite{qu2024connect}.

\paragraph{BirdCalls.}
We evaluate our method on the BirdCalls dataset using a pretrained EfficientNet-B0 \cite{tan2019efficientnet} encoder in Sections 4.2 and 4.3. Following \cite{gao2023out}, we first perform 10 epochs of linear probing (LP-FT \cite{kumar2022fine}) with the Adam optimizer, selecting the checkpoint with the best OOD validation performance to initialize the fine-tuning stage for 30 epochs.
Consistent with the protocol in \cite{qu2024connect}, we sample the linear probe learning rate from $10^{\text{Uniform}[-3,-2]}$, the fine-tuning learning rate from $10^{\text{Uniform}[-4,-3]}$, the augmentation probability from $\text{Uniform}[0.5, 0.9]$. The final hyperparameter setting is selected based on the best OOD validation performance. Reported results are averaged over 5 random seeds.  Other hyperparameters are the same as those of Gao et al. \cite{gao2023out}.

\paragraph{Galaxy10.}
We evaluate our method on the Galaxy10 DECaLS and Galaxy10 SDSS datasets with the ResNet18 \cite{he2016deep} backbone network in Sections 4.2 and 4.3. Specifically, we use the DECaLS for training and validation, and evaluate the model on the SDSS subset, where labels are mapped to DECaLS categories using a predefined label conversion rule since the labels in the two datasets are not exactly the same.
Our training pipeline employs a ResNet18 encoder initialized from a contrastively pre-trained checkpoint, followed by a linear classifier trained end-to-end. We first perform 10 epochs of linear probing (LP-FT \cite{kumar2022fine}) with the Adam optimizer, selecting the checkpoint with the best OOD validation performance to initialize the fine-tuning stage. The fine-tuning stage lasts for 15 epochs using the Adam optimizer with a learning rate of $10^{-4}$ and a batch size of 8. The augmentation probability is set to 0.5.

\paragraph{Benchmark Datasets.} And in Sections 4.2 and 4.3, we use the same ConvNet backbone as \cite{zhou2020deep, xu2023semantic, zhou2020learning, xu2023fourier} for Digits-DG, and ResNet18 \cite{he2016deep} for OfficeHome and PACS. Please find the details in the original paper and the released source code.

\paragraph{}
In D-GAP, the target image for mixing can be selected from either labeled or unlabeled data. For consistency and fair comparison with baseline methods \cite{gao2023out, qu2024connect}, we adopt a unified strategy across datasets. Specifically, for iWildCam, Camelyon17, and BirdCalls, we mix each source image with samples from other domains within the training set. For Galaxy10, since the training data only contains images from a single telescope, we randomly sample images from the unlabeled test set as mixing targets to generate cross-domain augmented images. For Digits-DG, OfficeHome, and PACS, we follow the target-source splitting as in \cite{xu2023semantic, xu2023fourier}. And the unlabeled target image for mixing can be selected from the held-out target domain during training.

\subsection{Baseline Methods}
For baseline methods on iWildCam, Camelyon17, BirdCalls dataset, we followed the settings and hyperparameters exactly in \cite{qu2024connect,gao2023out}. For the baseline methods for Digits-DG, OfficeHome, and PACS, we follow the settings in their original papers \cite{xu2023semantic, xu2023fourier, wong2025approximate, su2024enhanced}. And for Galaxy10 dataset, the settings are presented in Table \ref{settings}.

\section{Computation of Connectivity}
\label{appendix_connectivity}
To compute the average connectivity between two class-domain pairs $(c, d)$ and $(c', d')$, the process involves labeling all training examples from class $c$ and domain $d$ as 0, and those from class $c'$ and domain $d'$ as 1, while discarding examples from other classes or domains. 
Then, a ResNet50 model is trained using augmentations, stochastic gradient descent (SGD) with momentum, and a cosine learning rate schedule, maintaining consistent training parameters across all classes and domains. Specifically, we train binary classifiers from scratch to predict the class-domain pair of each input example. The test set is prepared similarly to the training set, and the model's performance on the test dataset provides an estimate of connectivity between the specified class-domain pairs, quantified by the test error rate. For iWildCam, we randomly select 15 class-domain pairs, while for Camelyon17, we use all class-domain pairs since it is a binary classification task. We train using the same hyperparameters described in \ref{Experimental Details} for 3,000 steps, applying early stopping based on validation accuracy. We will release the code of connectivity computing after the paper is published.

\section{Broader Impact}
This work studies dataset-agnostic augmentation for improving out-of-domain robustness in visual domain adaptation. Its potential positive impact lies in improving the reliability of machine learning models under distribution shifts, especially in scientific, ecological, and medical imaging scenarios where collecting labeled target-domain data is costly. By reducing the need for dataset-specific expert augmentation design, the proposed method may lower the barrier for deploying robust models in diverse real-world environments.

Potential negative impacts are limited but should not be ignored. Since the method improves model robustness under domain shifts, it could also be applied to sensitive visual recognition systems. In high-stakes domains such as medical imaging, improved OOD performance does not guarantee clinical reliability, and models should not be deployed without domain-specific validation, uncertainty estimation, and human expert oversight.

\newpage
\newpage
\section*{NeurIPS Paper Checklist}

\begin{enumerate}

\item {\bf Claims}
    \item[] Question: Do the main claims made in the abstract and introduction accurately reflect the paper's contributions and scope?
    \item[] Answer: \answerYes{} 
    \item[] Justification: The abstract and introduction clearly states the claims made, including the contributions made in the paper and important assumptions and limitations. 

\item {\bf Limitations}
    \item[] Question: Does the paper discuss the limitations of the work performed by the authors?
    \item[] Answer: \answerYes{} 
    \item[] Justification: Discussed in Section 5 Conclusion, Limitations and Future Work.

\item {\bf Theory assumptions and proofs}
    \item[] Question: For each theoretical result, does the paper provide the full set of assumptions and a complete (and correct) proof?
    \item[] Answer: \answerNA{} 
    \item[] Justification: It is an empirical paper that proposes a new method and justifies it through experimental results, intuitive explanations, and references to existing theoretical frameworks (e.g., feature decomposition and connectivity).

    \item {\bf Experimental result reproducibility}
    \item[] Question: Does the paper fully disclose all the information needed to reproduce the main experimental results of the paper to the extent that it affects the main claims and/or conclusions of the paper (regardless of whether the code and data are provided or not)?
    \item[] Answer: \answerYes{}  
    \item[] Justification: Experimental setups are reported in section Experiments and the Appendix. The code is provided in the supplementary material. And we will release the code after the paper is accepted.

\item {\bf Open access to data and code}
    \item[] Question: Does the paper provide open access to the data and code, with sufficient instructions to faithfully reproduce the main experimental results, as described in supplemental material?
    \item[] Answer: \answerYes{} 
    \item[] Justification: The code and the access to the datasets is provided in the supplementary material. All the code we used for the experiments will be public if the paper is accepted.

\item {\bf Experimental setting/details}
    \item[] Question: Does the paper specify all the training and test details (e.g., data splits, hyperparameters, how they were chosen, type of optimizer) necessary to understand the results?
    \item[] Answer: \answerYes{} 
    \item[] Justification: Some details are in Section Experiments. Other details are provided in our appendix and code in the supplemental material.

\item {\bf Experiment statistical significance}
    \item[] Question: Does the paper report error bars suitably and correctly defined or other appropriate information about the statistical significance of the experiments?
    \item[] Answer: \answerYes{} 
    \item[] Justification: We provide error bars in our experimental results and variance variance over multiple seeds in Appendix.

\item {\bf Experiments compute resources}
    \item[] Question: For each experiment, does the paper provide sufficient information on the computer resources (type of compute workers, memory, time of execution) needed to reproduce the experiments?
    \item[] Answer: \answerYes{} 
    \item[] Justification: This paper provides the compute resources in Section 4.1-Experimental Settings.
    
\item {\bf Code of ethics}
    \item[] Question: Does the research conducted in the paper conform, in every respect, with the NeurIPS Code of Ethics \url{https://neurips.cc/public/EthicsGuidelines}?
    \item[] Answer: \answerYes{} 
    \item[] Justification: The research conducted in the paper conforms, in every respect, with the NeurIPS Code of Ethics.

\item {\bf Broader impacts}
    \item[] Question: Does the paper discuss both potential positive societal impacts and negative societal impacts of the work performed?
    \item[] Answer: \answerYes{} 
    \item[] Justification: We discuss potential positive impacts in improving robust visual recognition for scientific, ecological, and medical imaging, as well as potential risks when the method is used in sensitive visual recognition or high-stakes decision systems (Appendix Sec. G).
    
\item {\bf Safeguards}
    \item[] Question: Does the paper describe safeguards that have been put in place for responsible release of data or models that have a high risk for misuse (e.g., pre-trained language models, image generators, or scraped datasets)?
    \item[] Answer: \answerNA{} 
    \item[] Justification: This paper poses no such risks.

\item {\bf Licenses for existing assets}
    \item[] Question: Are the creators or original owners of assets (e.g., code, data, models), used in the paper, properly credited and are the license and terms of use explicitly mentioned and properly respected?
    \item[] Answer: \answerYes{} 
    \item[] Justification: The original paper that produced the code package or dataset are cited.

\item {\bf New assets}
    \item[] Question: Are new assets introduced in the paper well documented and is the documentation provided alongside the assets?
    \item[] Answer: \answerYes{} 
    \item[] Justification: See supplementary material.

\item {\bf Crowdsourcing and research with human subjects}
    \item[] Question: For crowdsourcing experiments and research with human subjects, does the paper include the full text of instructions given to participants and screenshots, if applicable, as well as details about compensation (if any)? 
    \item[] Answer: \answerNA{} 
    \item[] Justification: This paper does not involve crowdsourcing nor research with human subjects.

\item {\bf Institutional review board (IRB) approvals or equivalent for research with human subjects}
    \item[] Question: Does the paper describe potential risks incurred by study participants, whether such risks were disclosed to the subjects, and whether Institutional Review Board (IRB) approvals (or an equivalent approval/review based on the requirements of your country or institution) were obtained?
    \item[] Answer: \answerNA{} 
    \item[] Justification: This paper does not involve crowdsourcing nor research with human subjects.

\item {\bf Declaration of LLM usage}
    \item[] Question: Does the paper describe the usage of LLMs if it is an important, original, or non-standard component of the core methods in this research? Note that if the LLM is used only for writing, editing, or formatting purposes and does \emph{not} impact the core methodology, scientific rigor, or originality of the research, declaration is not required.
    \item[] Answer: \answerNA{} 
    \item[] Justification: The core method development in this research does not involve LLMs as any important, original, or non-standard components.

\end{enumerate}

\end{document}